\documentclass{article}

\usepackage{arxiv}

\usepackage[utf8]{inputenc} 
\usepackage[T1]{fontenc}    
\usepackage{hyperref}       
\usepackage{url}            
\usepackage{booktabs}       
\usepackage{amsfonts}       
\usepackage{nicefrac}       
\usepackage{microtype}      
\usepackage{lipsum}
\usepackage{amsfonts,amssymb,amsbsy,textcomp,marvosym,amsmath,threeparttable,amsthm,subfigure,float,lastpage,lscape}
\usepackage{eurosym,mathrsfs,fancyhdr,CJK,multicol,graphics,indentfirst,color,bm,upgreek,booktabs,graphicx,multirow,warpcol}
\graphicspath{ {./images/} }

\title{PolicyEvolve: Evolving Programmatic Policies by LLMs for multi-player games via Population-Based Training}

\author{
 Mingrui Lv\textsuperscript{\#} \\
  School of Computer Science \\
  Sichuan Normal University\\
  Chengdu, Sichuan, China \\
  \texttt{2023110133@stu.sicnu.edu.cn} \\
  \And
  Hangzhi Liu\textsuperscript{\#} \\
  School of Computer Science \\
  Sichuan Normal University\\
  Chengdu, Sichuan, China \\
  \texttt{liuhanzhi49@outlook.com} \\
  \And
  Zhi Luo \\
  School of Computer Science \\
  Sichuan Normal University\\
  Chengdu, Sichuan, China \\
  \texttt{sixi225307@gmail.com} \\
  \AND
  Hongjie Zhang* \\
  School of Computer Science \\
  Sichuan Normal University\\
  Chengdu, Sichuan, China \\
  \texttt{zhanghongjie@sicnu.edu.cn} \\
  \And
  Jie Ou \\
  School of Information and Software Engineering \\
  University of Electronic Science and Technology of China \\
  Chengdu, Sichuan, China \\
  \texttt{oujieww6@gmail.com} \\
}

\begin{document}
\footnotetext{\# These authors contributed equally to this work.}
\footnotetext{* Corresponding author}
\maketitle
\begin{abstract}
Multi-agent reinforcement learning (MARL) has achieved significant progress in solving complex multi-player games through self-play. However, training effective adversarial policies requires millions of experience samples and substantial computational resources. Moreover, these policies lack interpretability, hindering their practical deployment. Recently, researchers have successfully leveraged Large Language Models (LLMs) to generate programmatic policies for single-agent tasks, transforming neural network-based policies into interpretable rule-based code with high execution efficiency. Inspired by this, we propose PolicyEvolve, a general framework for generating programmatic policies in multi-player games. PolicyEvolve significantly reduces reliance on manually crafted policy code, achieving high-performance policies with minimal environmental interactions. The framework comprises four modules: \textbf{Global Pool}, \textbf{Local Pool}, \textbf{Policy Planner}, and \textbf{Trajectory Critic}. The Global Pool preserves elite policies accumulated during iterative training. The Local Pool stores temporary policies for the current iteration; only sufficiently high-performing policies from this pool are promoted to the Global Pool. The Policy Planner serves as the core policy generation module. It samples the top three policies from the Global Pool, generates an initial policy for the current iteration based on environmental information, and refines this policy using feedback from the Trajectory Critic. Refined policies are then deposited into the Local Pool. This iterative process continues until the policy achieves a sufficiently high average win rate against the Global Pool, at which point it is integrated into the Global Pool. The Trajectory Critic analyzes interaction data from the current policy, identifies vulnerabilities, and proposes directional improvements to guide the Policy Planner in generating enhanced policies. We conduct extensive experiments using various LLMs on multi-player game tasks, comparing PolicyEvolve against state-of-the-art prompt-based techniques. Results demonstrate that PolicyEvolve significantly outperforms baseline methods in policy quality.
\end{abstract}


\section{Introduction}
Multi-agent reinforcement learning has achieved breakthrough progress in collaborative and swarm control systems, such as robotics\cite{kober2013reinforcement,rusu2017sim}, games\cite{kurach2020google,li2025comprehensive,subramanian2023robustness}, autonomous driving\cite{zhang2024multi,kiran2021deep} and social problems\cite{zhang2024equilibrium,baker2020emergent,mi2024taxai}. However, current MARL approaches still face core challenges. The decision-making mechanisms based on deep neural networks exhibit a "black-box" nature, rendering system behaviors difficult for human users to fully comprehend and trust. Furthermore, when multi-agent systems encounter unforeseen dynamic environments, their performance degrades significantly, revealing limitations in generalization capability. Recent studies emphasize that enhancing interpretability and environmental adaptability represents a critical direction for advancing real-world applications of this technology.

Recent advances in programmatic reinforcement learning (PRL) have established it as a promising approach within interpretable reinforcement learning. Unlike traditional deep reinforcement learning, which employs neural networks to represent state-action mappings, PRL utilizes rule-based code to encode policies. The logical rigor and inherent interpretability of code ensure policy safety. The rapid evolution of large language models (LLMs) has revolutionized code generation capabilities, making LLM-assisted generation of programmatic policies feasible. Deng et al. introduced the LLM-SMAC framework, where agents leverage LLMs to generate decision tree code by providing task descriptions. The resulting rule-based code achieved near-perfect win rates on both simple and complex StarCraft Multi-Agent Challenge (SMAC) tasks while ensuring policy safety and interpretability\cite{deng2024new}. For autonomous driving scenarios, Zeng et al. proposed the ADRD framework, employing LLMs to generate RULE-BASED DECISION code, which demonstrated superior policy quality and execution efficiency compared to conventional algorithms\cite{zeng2025adrd}. Sadmine et al. developed the LS-LLM algorithm, utilizing LLMs' code generation capabilities to write computer programs that accelerate programmatic policy synthesis. They specifically used LLM-generated policies solely as an initial population, evolving advanced strategies through heuristic search algorithms\cite{sadmine2024language}. Liu et al. presented the LLM-GS framework, designing a Pythonic-DSL policy representation. This approach first generates Python code, which is then converted into domain-specific languages (DSLs) to mitigate LLM hallucination issues regarding DSLs. Concurrently, LLM-GS employs a Scheduled Hill Climbing algorithm to enable efficient exploration within the code space, thereby enhancing policy quality\cite{liusynthesizing}.

\begin{figure}[!ht]
\centering
\centerline{\includegraphics[width=\textwidth]{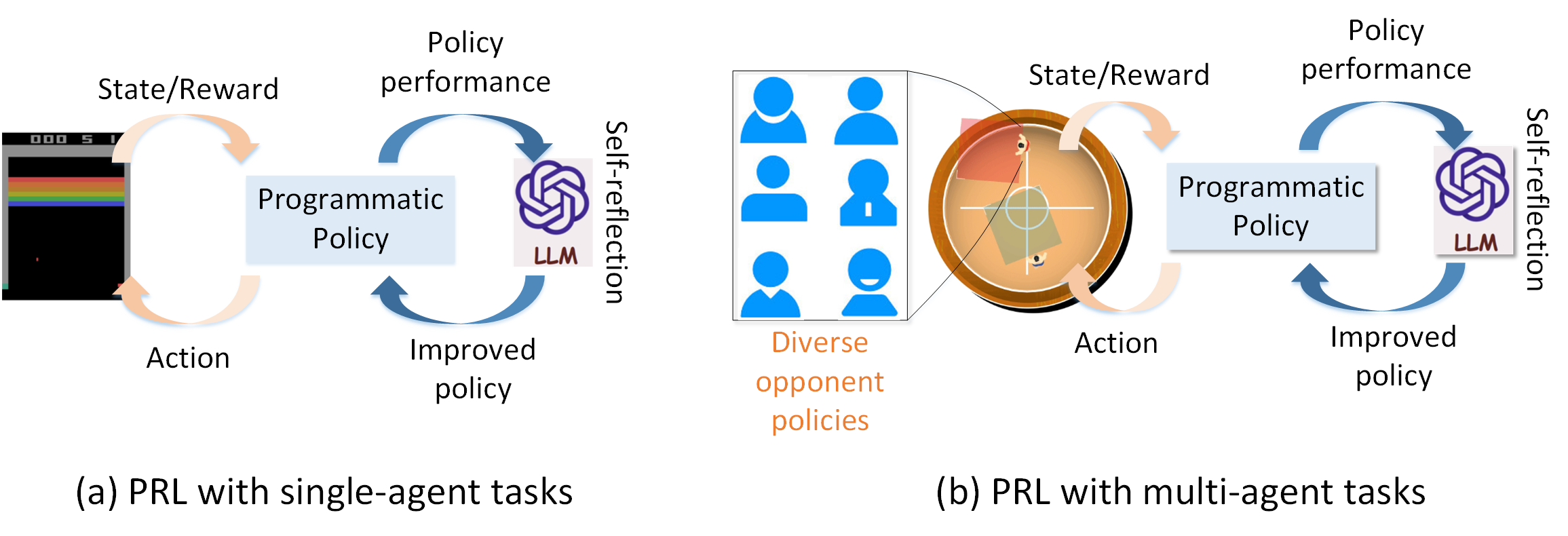}}
\caption{The differences in using PRL to solve single-agent versus multi-agent tasks.}
\label{fig:s&m}
\end{figure}

However, the aforementioned approaches predominantly address single-agent tasks characterized by static environments devoid of adversarial interference, rendering policy learning comparatively straightforward. In multi-agent scenarios, the core challenge lies in the necessity for continuous policy adaptation in response to evolving opponent policies, which inherently creates environmental instability (as shown in Fig.~\ref{fig:s&m}). We aim to develop policy code capable of autonomous evolution to maintain efficacy against diverse adversaries. To this end, we introduce \textbf{PolicyEvolve}—a general-purpose framework for multi-agent tasks that enables policy generation and improvement without human intervention. The core innovation involves maintaining a global policy pool preserving elite policies accumulated during iterative training, alongside a local pool storing temporary policies for the current iteration. A Policy Planner generates candidate policies for the current iteration based on environmental information and refines them using feedback from the Trajectory Critic, which analyzes interaction data to identify vulnerabilities and propose directional improvements. Refined policies are deposited into the Local Pool, and this iterative cycle persists until the policy achieves a sufficiently high average win rate against the Global Pool, at which point it is integrated into the Global Pool, thereby systematically evolving policies capable of adapting to dynamic multi-agent environments through population-based self-play against the global policy pool. We conducted extensive experiments on multi-player zero-sum games, validated across diverse open-source and proprietary large language models (LLMs). Notably, PolicyEvolve generates a high-quality white-box programmatic policy model through minimal environmental exploration. Our framework demonstrates consistent strategy evolution, progressively improving the ELO rating of the policies. Comparative evaluations against prompt-based baselines reveal that our framework yields policies achieving the highest average win rate against diverse opponent policies.

To summarize, we make the following contributions:

1. We introduce PolicyEvolve, the first programmatic reinforcement learning framework specifically designed for multi-agent tasks, which autonomously evolves policies to consistently enhance policy quality.

2. To improve policy robustness, we design Global and Local policy pools within PolicyEvolve, trained through a Population-Based Training approach.

3. Extensive experiments across diverse LLMs on multi-agent tasks demonstrate that our framework achieves significantly superior sample efficiency and policy quality compared to prompt-based baseline algorithms.

\section{Methodology}
\subsection{Overview}
\begin{figure}[!ht]
\centering
\centerline{\includegraphics[width=\textwidth]{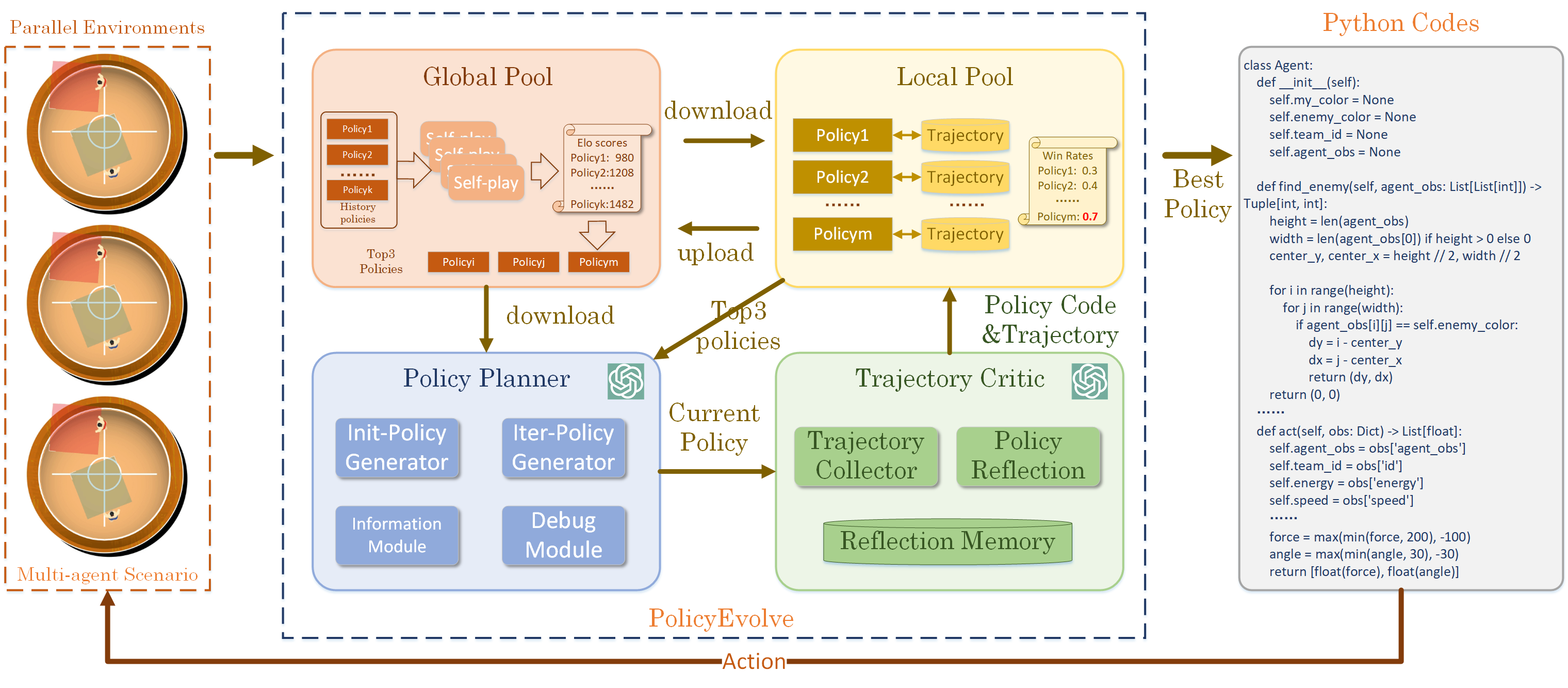}}
\caption{The high-level overview of the \textit{PolicyEvolve} framework.}
\label{fig:overview}
\end{figure}

Figure \ref{fig:overview} shows a high-level overview of the \textit{PolicyEvolve} framework. Specifically, PolicyEvolve comprises four modules: Global Pool, Local Pool, Policy Planner, and Trajectory Critic. Centered around population-based training, the Global Pool preserves elite populations. It continuously samples multiple policies for adversarial evaluations, ranking them via an ELO scoring mechanism to guide subsequent opponent sampling and new policy iterations. The Policy Planner leverages LLMs to generate new policy code. These policies engage in adversarial play against the Global Pool's population, analyzing opponent tactics and self-deficiencies to iteratively evolve superior polcies that join the Global Pool. This cycle repeats until policy improvements plateau. Given potential code bugs in LLM-generated outputs, the Planner iteratively debugs the code. Debugged policies may still lack sufficient strength; thus, they undergo win-rate evaluation against randomly sampled opponents from the Global Pool. Polcies failing to meet strength thresholds are excluded from the Global Pool. To address this, we introduce the Local Pool, which stores historical policies from the current iteration. Current policies and adversarial interaction data are archived in the Local Pool. To diagnose policy failures, the Trajectory Critic module employs LLMs' reasoning capabilities to analyze policy code and interaction trajectories, identifying failure root causes and generating summaries. It then provides improvement suggestions through policy reflection. The Policy Planner revises code based on these suggestions and revalidates policy efficacy. This iterative process persists until the policy achieves a win rate exceeding 60\%, indicating sufficient quality for promotion to the Global Pool. At this stage, the Local Pool is reset to await the next iteration. Through continuous algorithmic iterations, the Global Pool progressively generates increasingly superior policies.

\subsection{Global Pool}
The core of Population-Based Training lies in maintaining a population of policies that are both high-performing and diverse. Through continuous adversarial interactions within the population, new superior policies progressively evolve. Our designed Global Pool preserves elite historical policies throughout the training framework’s iterations. Particularly, when a new policy achieves a 60\% win rate against policies in the Global Pool, it is admitted into the Global Pool for evaluation. Policies within the Global Pool engage in pairwise matches, with their rankings determined by ELO scores to select high-quality candidates for subsequent evolution. The Global Pool randomly samples $n$ policies to serve as agents in the operational environment for policy evaluation. Each strategy’s win rate is computed as the average outcome over $t$ evaluations. The ELO scores of these $n$ policies are updated according to the ELO calculation formula, as shown in Eq.~\ref{equ:elo}. The fundamental principle of the ELO system is that each player possesses a rating, which is adjusted post-match based on outcomes. Winners “gain” points from losers, while losers “lose” points, with the magnitude of change contingent upon the pre-match rating disparity. Specifically, each newly admitted policy to the Global Pool is assigned an initial score (e.g., 1200).

\begin{equation}
\begin{split}
& E_A=\frac{1}{1+10^{(R_B-R_A)/400}}, \\
& R_A'=R_A+K\cdot(S_A-E_A) \label{equ:elo}
\end{split}
\end{equation}
Before the match begins, the system calculates the expected score for each player based on their current ratings, where $R_A$ represents player A's current rating, $R_B$ represents player B's current rating, and $E_A$ represents player A's expected win rate against player B. After the match executes, each player's actual score (S) is determined as follows: win: $S=1$, draw: $S=0.5$, loss: $S=0$. R' represents the new rating calculated for each player based on the difference between the actual score and the expected score. $K$ is an adjustment coefficient that controls the magnitude of rating changes, typically set according to match importance, player experience, and other factors (for example, K=32). As the number of policies in the Global Pool increases, the frequency of self-play correspondingly rises. To rapidly evaluate the ELO score of each policy, we developed a parallel evaluation algorithm. Upon selecting $n$ policies, this algorithm initiates $t$ evaluation processes that conduct assessments concurrently, returning win-loss outcomes.

\subsection{Local Pool}
\begin{figure}[!ht]
\centering
\centerline{\includegraphics[width=\textwidth]{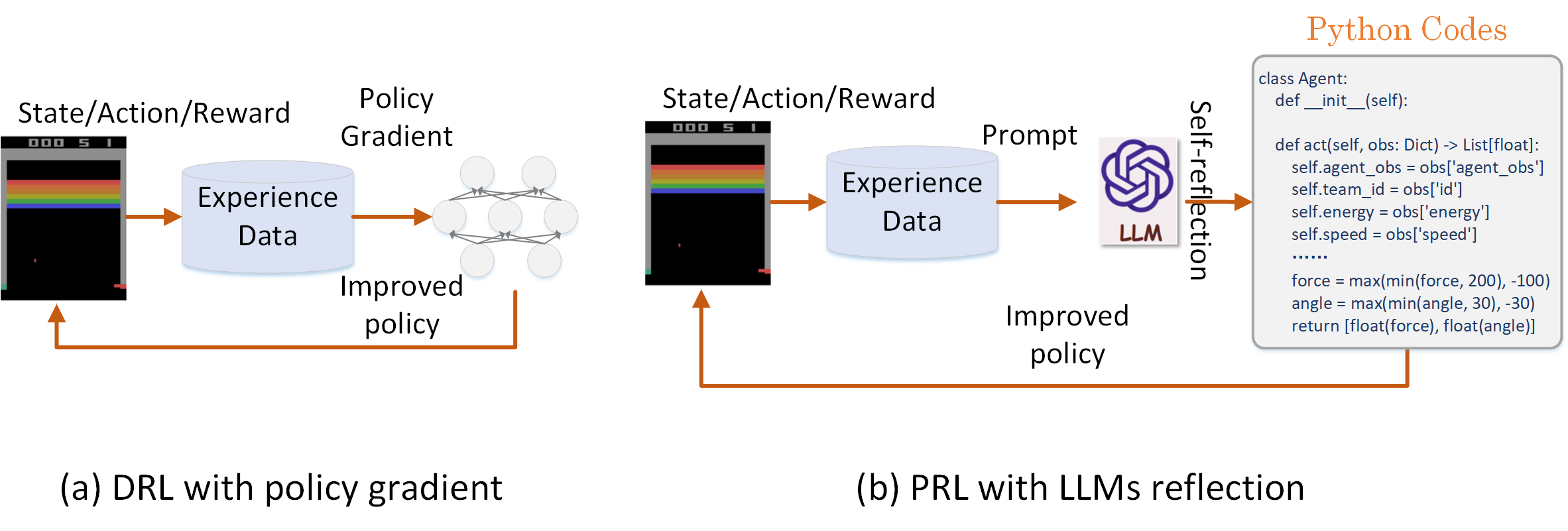}}
\caption{Comparison of DRL and PRL in utilizing empirical data for policy updates.}
\label{fig:drl&prl}
\end{figure}
Policies generated by LLMs may not be optimal, potentially exhibiting lower win rates against those in the Global Pool. Consequently, we employ local multi-round iterations to continuously refine the LLM-generated code, integrating it into the Global Pool only upon achieving a sufficiently high win rate. This process is analogous to policy gradients in deep reinforcement learning (DRL), as depicted in Fig.~\ref{fig:drl&prl}, where the current policy competes against opponents to collect (state, action, reward) experience data. DRL computes policy gradients from this data to update neural network parameters and enhance policy quality. Similarly, LLMs directly analyze success/failure reasons within the experience data to modify policy code and improve quality. Both approaches fundamentally adjust policies based on empirical data, though DRL operates with greater granularity. In DRL, policy quality may temporarily degrade after parameter updates due to suboptimal experiences acquired during agent exploration, leading to unfavorable gradient directions; however, stable improvement emerges over multiple iterations. Likewise, updating policy code via LLMs can cause quality degradation, primarily attributed to limitations in the experience data and inherent hallucination issues of LLMs. Hence, multi-round iterative updates are essential to ensure stable quality enhancement. Suboptimally refined policies during this process are stored in a temporary repository designated as the Local Pool. This Local Pool not only preserves policy code but also retains corresponding experience data from battles against the Global Pool, functioning analogously to the experience replay buffer in DRL. These experience data are subsequently employed by the Trajectory Critic for policy reflection and quality improvement. Additionally, the Local Pool also stores the average win rate of each policy, which is used by the Policy Planner for policy selection.

\subsection{Policy Planner}
The Policy Planner serves as the core component for generating policy code using LLMs, comprising four submodules: Init-Policy Generator, Iter-Policy Generator, Information Module, and Debug Module. Specifically, the Init-Policy Generator produces the initial policy for the current iteration round; the Iter-Policy Generator generates improved policies based on those stored in the Local Pool; the Information Module provides task environment information described by users in natural language, serving as the foundational information for policy generation; and the Debug Module is responsible for debugging the code generated by LLMs to ensure its executability.

\begin{figure}[!ht]
\centering
\centerline{\includegraphics[width=\textwidth]{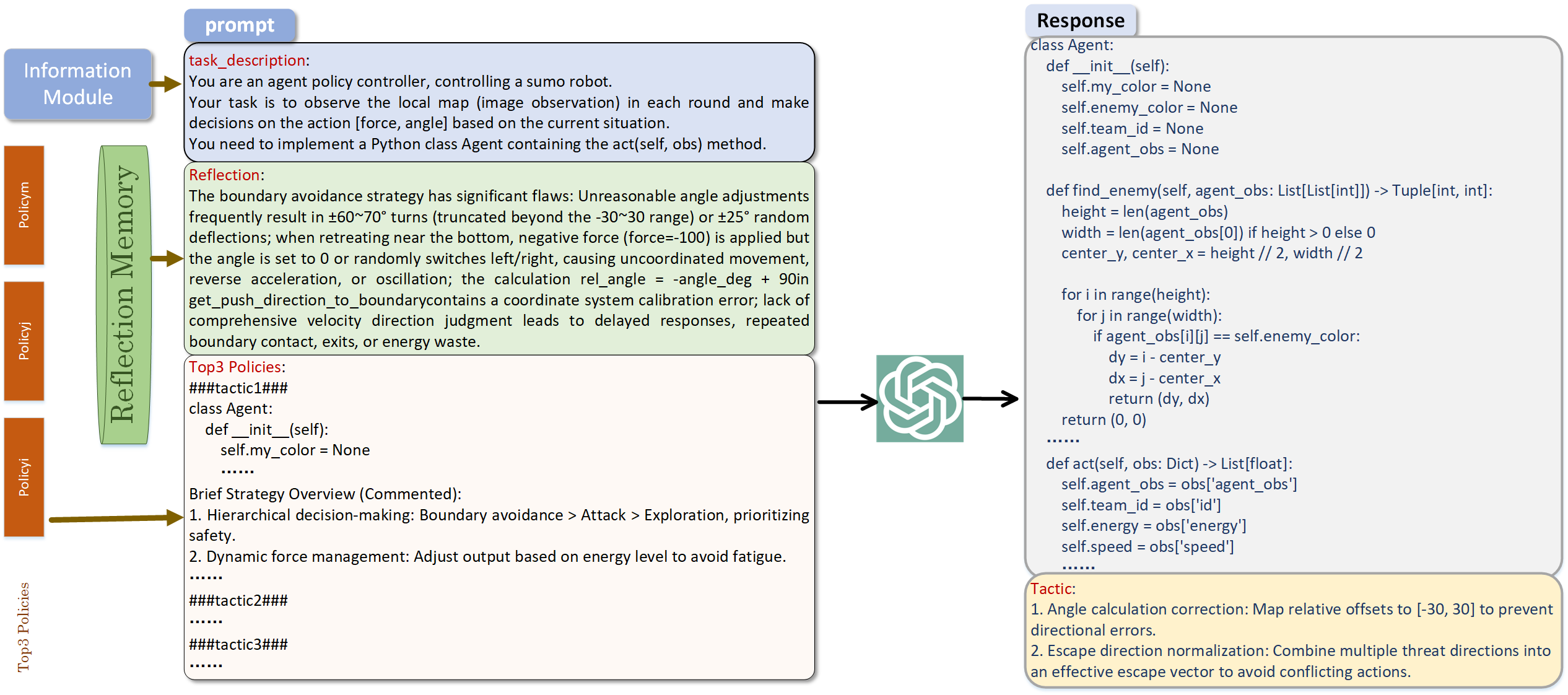}}
\caption{Initial policy generation process and prompts.}
\label{fig:init-g}
\end{figure}

\paragraph{Init-Policy Generator}
The quality of the initial policy directly impacts the entire evolutionary process. Therefore, our initial policy consists of two components. First, upon the initial launch of PolicyEvolve, the Global Pool contains only a single random policy (i.e., actions generated randomly, independent of environmental states), and LLMs generate policy code solely based on the current environmental description, where the level of detail in the task description determines the quality of the initial policy. Second, after multiple iterations, the Global Pool already contains high-quality policy code; at this stage, we select the three highest-scoring policy codes from the Global Pool based on ELO scores and place them into the Local Pool as seed policies. The Policy Planner retrieves these three highest-scoring policy codes from the Local Pool and obtains corresponding policy explanations for each from the Trajectory Critic to facilitate understanding of the policy code. Simultaneously, it acquires the environmental description from the Information Module and retrieves long-term reflection memories of historical policies from the Trajectory Critic's reflection memory pool. The Trajectory Critic then concatenates these elements to form a complete context prompt for code generation, from which LLMs generate the initial policy code, as illustrated in Fig.~\ref{fig:init-g}. The policy code will be executed in a Python environment, and compilation errors will be captured. The Traceback and code will be sent to Debug Module to fix the code bugs, ultimately generating a runnable initial policy code.

\paragraph{Iter-Policy Generator}
The policy generated by the Init-Policy Generator may underperform against policies in the Global Pool, necessitating multiple iterations within the Iter-Policy Generator until the policy's win rate exceeds 60\%, at which point the current policy code is uploaded to the Global Pool. The key to policy enhancement lies in the in-depth analysis of current policy code and battle experience data: when battling against the Global Pool, both the causes of failures and the key factors for successes must be identified and recorded for subsequent policy improvement. This analysis of experience data is conducted by the Trajectory Critic, from which the Iter-Policy Generator extracts the analytical conclusions. These conclusions are then concatenated with environmental information and the current code to form a context prompt, enabling LLMs to regenerate enhanced policy code, with the specific prompt structure detailed below:

\textit{Environmental Information: [promote](\#promote)}

\textit{Old Code: [old\_code](\#old\_code)}

\textit{Reflection Results/ Code Error: reflex (previous code error and reflection; refer to [reflexion](\#reflexion) for output format)}

\textit{[!important] Please generate a new strategy. The code you produce must not duplicate the old code. Endeavour to ensure opponents go out of bounds while remaining in bounds yourself. Provide directly executable code.}

\paragraph{Information Module}
The task description should be provided by users with as much detail as possible, including task objectives/rewards, state content/explanations, and action space/explanations; heuristic prompts to guide policy formulation may also be included. Below is the environmental description prompt we designed for the \textit{wrestle} task:

\textit{You are an intelligent agent strategy controller, operating a sumo robot.}

\textit{Your task is to observe the local map (image obs) each turn and decide on an action `[force, angle]` based on the current situation.}

\textit{- The input parameter 'obs' is a dictionary representing an agent's observation information.}

\textit{- Each dictionary contains the following fields:}

   \textit{- 'agent\_obs':A two-dimensional list, where elements are colour-coded integers, accessed via integer 
   indices, such as `agent\_obs[i][j]`}
   
   \textit{- 'id':Agent team identifier, with values 'team\_0' or 'team\_1'.}
   
   \textit{- 'energy':The current energy value of the agent is of type float. For example:100.0}
   
   \textit{- 'speed':The current velocity vector of the agent (x, y), for example:[2.5, 1.8].}
   
\textit{- It is known that the colour code for `team\_0` is 10, and the colour code for `team\_1` is 8.}

\textit{[Environmental Monitoring obs]}

\textit{- `agent\_obs` is a two-dimensional list whose elements are integer colour codes. It must be accessed using integer indices, such as `agent\_obs[i][j]`.}

     \textit{1: green      → Touching the boundary line results in a loss}
     
     \textit{2: sky blue   → Auxiliary line}
     
     \textit{4: grey       → Central Security Zone}
     
     \textit{10: red       → `team\_0` Agent colour}
     
     \textit{8: blue       → `team\_1` Agent colour}
     
\textit{[Physical laws]}

\textit{- Actions are converted into acceleration vectors, with the environment updating velocity and position at each step.}

\textit{- Speed decays by 0.98 per step, subject to a maximum speed limit.}

\textit{- Touching the green boundary (Code 1) constitutes an out-of-bounds violation, deemed a failure, and must be avoided.}

\textit{[Physical Stamina Mechanism]}

\textit{- The agent possesses its own energy, with the energy expended per step being proportional to both the applied driving force and the displacement.}

\textit{- The agent's energy recovers at a constant rate. Should energy deplete to zero, the agent becomes fatigued, rendering it unable to exert force and incapable of movement.}

\textit{- Efforts should be planned with reasonable intensity and pacing to avoid prolonged periods of high-intensity output.}

\paragraph{Debug Module}
The code generated by LLMs may contain compilation errors due to hallucination issues, such as undefined variables or division-by-zero errors. We execute the generated code within a Python virtual machine and capture the compiler's traceback, then directly feed both the traceback and the source code to the LLMs, instructing them to fix the code bugs. This process may be repeated multiple times until all code bugs are resolved. Specifically, the debug prompt designed in this paper is as follows:

\textit{You are an expert code debugger and software engineer. Your task is to analyze the provided Python code and its accompanying compilation traceback error, then generate a fixed version of the code that resolves the error completely.}

\textit{Traceback Error: ......}

\textit{Original Code: ......}

\textit{Output Format: ......}

\textit{Fixed Code:}

\subsection{Trajectory Critic}
Policy reflection is crucial for enhancing policy quality, requiring analysis of both the policy code itself and adversarial experience data to identify the causes of success or failure and guide directions for policy improvement, analogous to policy gradient information in DRL. The Trajectory Critic comprises three submodules: Trajectory Collector, Policy Reflection, and Reflection Memory.

\paragraph{Trajectory Collector}
This module engages the current policy in adversarial interactions against policies from the Global Pool to collect experience data. Unlike DRL approaches, these experience data require semantic annotation for analysis by LLMs. Annotating each data entry individually would consume excessive tokens, preventing LLMs from capturing critical information. Therefore, we store experience data in the following \textbf{JSON} format, merging similar data entries to reduce token consumption.

\textit{```json}
\textit{\{}
\textit{  "an": "20250830\_114344",}
\textit{  "rs": [0.0,0.8,1.0],}
\textit{  "fr": [0.0,1.0,1.0],}
\textit{  "af": [[80.0,80.0,],[80.0,80.0,],[80.0,80.0,]],}
\textit{  "aa": [[20.0,20.0,],[20.0,20.0,],[16.83,20.0,]],}
\textit{  "of": [[80.0,80.0,],[139.23,114.42,],[51.82,100.59,]],}
\textit{  "oa": [[30.0,-30.0,],[30.0,30.0,],[-8.10,-18.57,]],}
\textit{  "d2b": [[19.72,17.0,],[19.72,17.0,],[20.12,17.49,]],}
\textit{  "rw": [[0,0,],[0,0,],[0,0,]]}
\textit{\}}
\textit{```}

In the design of selecting opponent policies from the Global Pool, we employ stochastic sampling based on ELO scores. Each policy's strength is represented by its ELO score, with higher scores indicating greater strength. A softmax-based approach is applied to select $k$ opponent policies according to their respective ELO scores, initiating matches as shown in Eq.~\ref{equ:softelo}. The purpose is to ensure comprehensive evaluation by selecting $k$ opponents via softmax, thereby requiring new policies to maintain high win rates against diverse opponent types to demonstrate their generality and effectiveness.

\begin{equation}
p_A=\frac{e^{R_A}}{\sum_k{e^{R_k}}}\label{equ:softelo}
\end{equation}

\paragraph{Policy Reflection}
After obtaining experience data, we analyze it to identify critical factors for success or failure. Specifically, we require LLMs to concisely summarize the agent's overall performance and key issues during matches, list specific code logic errors with location information for each error, and propose concrete and actionable improvement suggestions including code examples. The prompt of Policy Reflection is:

\textit{You are now required to analyse the agent's strategy flaws based on the match log and generate improvement proposals. Below are the key log information and analysis requirements:}

\textit{[Log Field Descriptions]}

\textit{- an: Current Agent Name (Generated from the filename, such as ‘20250801\_143000’)}

\textit{- rs: Win rate against opponent AI agents}

\textit{- fr: Multi-game final reward (float array)}

\textit{- af: Force applied by the agent per move per game (two-dimensional array)}

\textit{- aa: Agent's perspective per move per game (two-dimensional array)}

\textit{- of / oa: Opponent's movement data (two-dimensional array)}

\textit{- d2b: Centre point to boundary minimum distance (two-dimensional array, unit: pixels)}

\textit{- rw: Step Reward Sequence (two-dimensional array)}

\textit{[Output Requirements][!important]}

\textit{- The output code must ultimately display the following content and include a comment}

\textit{  \#Reflection: A concise summary of the AI agent's overall performance and key issues during the match.}

\textit{  \#1. [Specific Description of Strategic Flaw 1]}

\textit{  \#2. [Specific Description of Strategic Flaw 2]}

\textit{  \# // ... Entries may be added or removed as appropriate ...}

\textit{  \#Code error: List specific code logic errors, each error must include location information.}

\textit{  \#1. [The precise location and description of Error 1]}

\textit{  \#2. [The precise location and description of Error 2]}

\textit{\# // ... Entries may be added or removed as appropriate ...}

\textit{  \#Improvement Recommendations: For each error, propose specific, actionable improvement plans, including code examples}

\textit{  \#1. [Specific Improvement Plan for Error 1]}

\textit{  \#2. [Specific Improvement Plan for Error 2]}

\textit{\# // ... Entries may be added or removed as appropriate ...}

\textit{  The output must strictly adhere to the above three-part structure, no additional content should be added.}

The results after reflection are stored in the Reflection Memory and are used by the Iter-Policy Generator to improve the policy, ultimately generating an enhanced policy code that is saved in the Local Pool.

\paragraph{Reflection Memory}
\begin{figure}[!ht]
\centering
\centerline{\includegraphics[width=\textwidth]{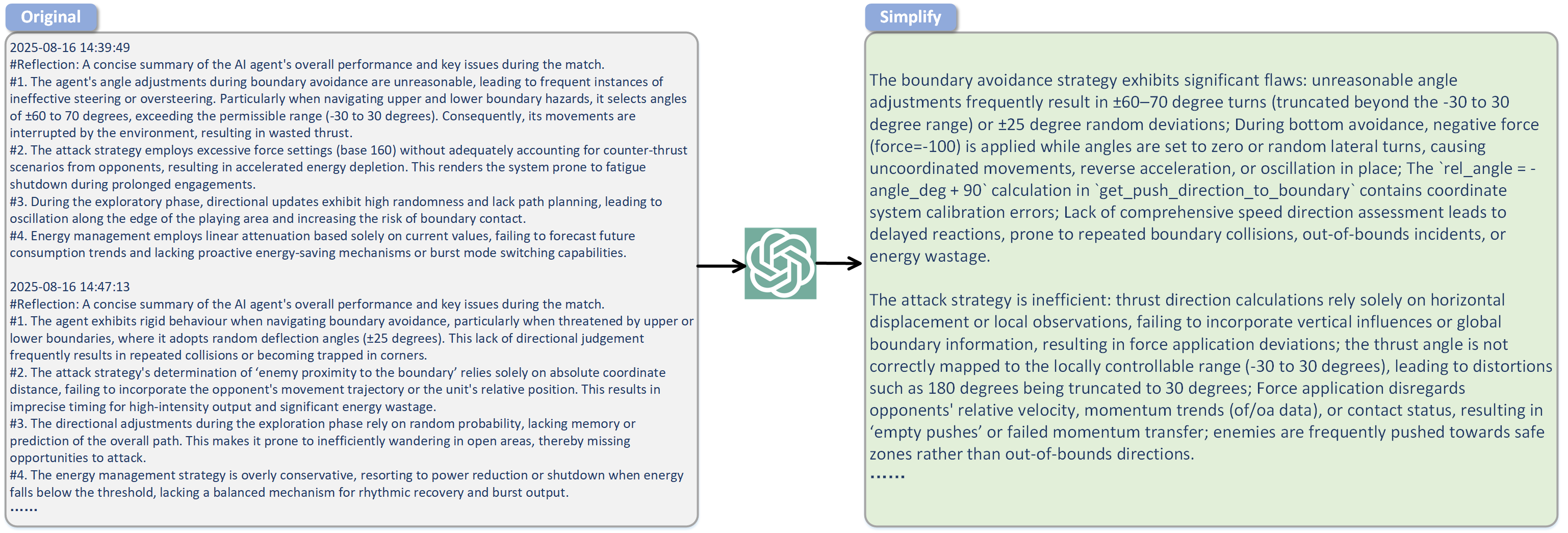}}
\caption{Reflection Memory structure for the wrestle task before and after simplification.}
\label{fig:reflectmem}
\end{figure}
Upon invoking LLMs for multiple iterations, the Reflection Memory becomes excessively large, leading to substantial token consumption and degradation in code quality. We address this by invoking LLMs to summarize and simplify the Reflection Memory content, eliminating redundant descriptions and repetitive content while using concise language to encapsulate key information. The prompt for simplifying Reflection Memory is as shown below:

\textit{Note:}
\textit{Please perform a lossless semantic simplification and consolidation of the memory\_pool (complete Markdown text) above, returning the final result **exclusively as plain Markdown text**: each merged question/suggestion should occupy a single line (exclude timestamps, JSON, code blocks, and additional explanatory notes). Strictly adhere to the following rules.}
        
\textit{Procedure (to be strictly adhered to):}

\textit{1. Split and delete timestamps: \#}

\textit{2. Extracting ‘semantic units’: \#}

\textit{3. Semantic Merging Rules (Concretisation of 'Merge when meanings are broadly similar'): \#}

\textit{4. Conflict and Duplicate Handling: \#}

\textit{5. Authenticity and Formatting Requirements: \#}

\textit{6. Output Requirements (Strictly Mandatory): \#}

\textit{7. Example Merge Style (for demonstration purposes only; not intended as output): \#}

\textit{Please now process memory\_pool according to the aforementioned rules, returning only the final, pure Markdown text (each line containing a single merged piece of information, sorted in descending order by importance/frequency). Thank you.}

The comparison of the Reflection Memory before and after simplification for the wrestle task is shown in Fig.~\ref{fig:reflectmem}. Before simplification, each iteration had its own policy reflections, but these contained duplicate or similar experiences that could be abstracted and summarized.

\section{Experiments}
We implement the PolicyEvolve framework using three large language models—Hunyuan, Qwen, and DeepSeek—and conducts extensive experiments on the \textit{Wrestle} task provided by the Chinese Academy of Sciences' JIDI platform. Comparisons against state-of-the-art prompt-based code generation techniques demonstrate that the policy code generated by our framework achieves the highest win rate and robustness. Furthermore, comprehensive ablation studies analyze the design advancements of the PolicyEvolve framework, validating the benefits contributed by each improved component.

\subsection{Experiments Setup}
\begin{figure}[!ht]
\centering
\centerline{\includegraphics[width=5in]{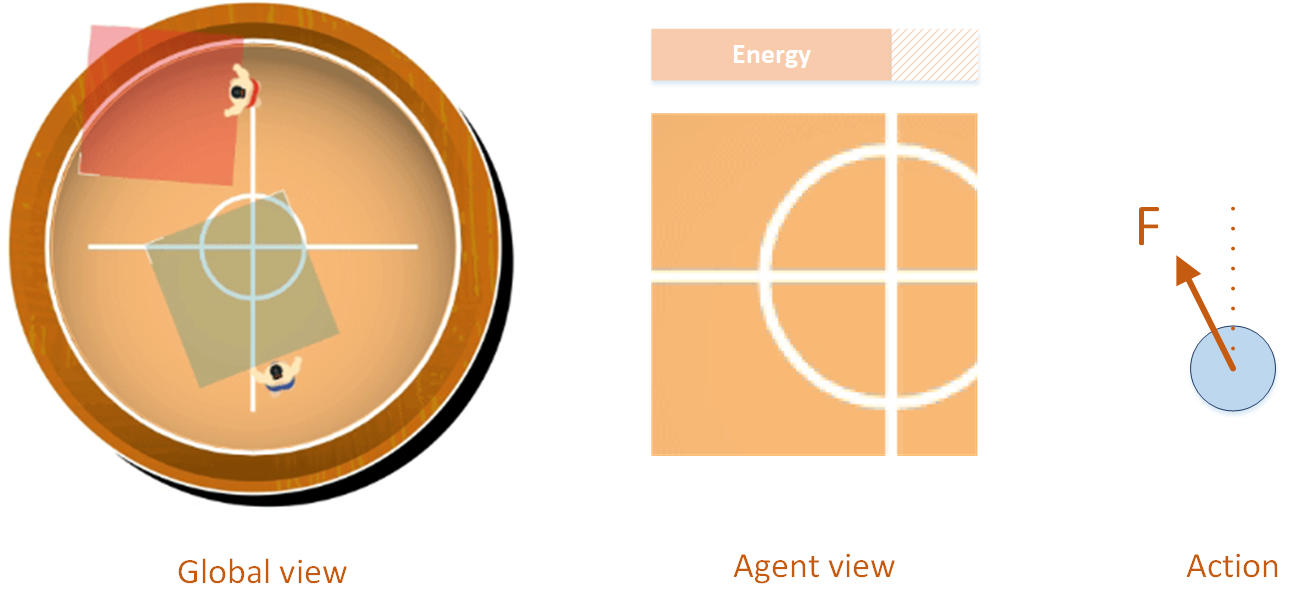}}
\caption{The global view, agent's local view, and action space for the wrestle task.}
\label{fig:wrestle}
\end{figure}

\paragraph{Task}
Wrestle is a two-player adversarial task provided by the Chinese Academy of Sciences, with its environment visualization and action space depicted in Fig.~\ref{fig:wrestle}. The game involves two parties, each controlling an elastic sphere agent of identical mass and radius positioned at opposite ends of a circular arena. Upon commencement, agents may move freely within the circular boundary but lose upon touching the perimeter, resulting in disqualification. Agents can collide with each other, experiencing velocity reduction proportional to the sphere's friction coefficient. Each agent possesses an energy reserve depleted proportionally to applied driving force and displacement per step; energy recovers at a fixed rate, and depletion to zero induces fatigue, disabling force application. The environment terminates when one agent falls off the ring or reaches the maximum 500-step limit.

Observation is a dictionary containing keys "obs" and "controlled\_player\_index". The "obs" value is a dictionary comprising a 40x40 matrix and supplementary game-related data; the matrix captures the agent's forward-facing field of view, revealing walls, ground markers, and other agents within range. The "controlled\_player\_index" value specifies the identifier of the controlled agent.
The action space is a list of length n\_action\_dim=2, where each element is a Gym Box class: [Box(-100.0, 200.0, (1,), float32), Box(-30.0, 30.0, (1,), float32)], representing applied force and steering angle respectively.

Reward: Knocking the opponent off the ring yields 100 points; otherwise, 0 points are awarded.

\paragraph{Baselines}
To verify the effect of PolicyEvolve on multi-player games, we compare to algorithms based on prompt engineering included \textit{Naive}, \textit{CoT}~\cite{wei2022chain} and \textit{React}~\cite{yao2023react}. Specifically, we compare all policy codes with \textit{Random} action. \textit{Random} ignores environmental state, generating force and steering angle actions directly via random numbers. \textit{Naive} feeds environmental descriptions, action space, and task objectives to LLMs through prompting to generate policy code, subsequently refined by our Debug module to ensure executability. \textit{Naive} only possesses environmental information without knowledge of opponent policies. \textit{CoT} guides LLMs via prompting to deeply consider constructing robust policy code and potential opponent strategies, generating policies that also undergo Debug module refinement for executability. \textit{React} pits LLM-generated policy code against Random policies, refining the policy based on empirical data, with each iteration's code corrected by the Debug module.

\paragraph{Metrics}
We evaluate policy quality using win rate (WR) and ELO score (as shown in Eq.~\ref{equ:elo}). The win rate is defined as follows: 
\begin{equation}
WR=\frac{\sum_L{s}}{L}\label{equ:succ}
\end{equation}
where $s$ takes values of -1, 0, and 1, representing loss, draw, and win in a match, respectively. Due to the randomness in opponent policies and the need to switch sides, $L$ random matches are played, with $L=10$ in this work.

\subsection{Main Results}
\begin{figure}[!ht]
\centering
\centerline{\includegraphics[width=\textwidth]{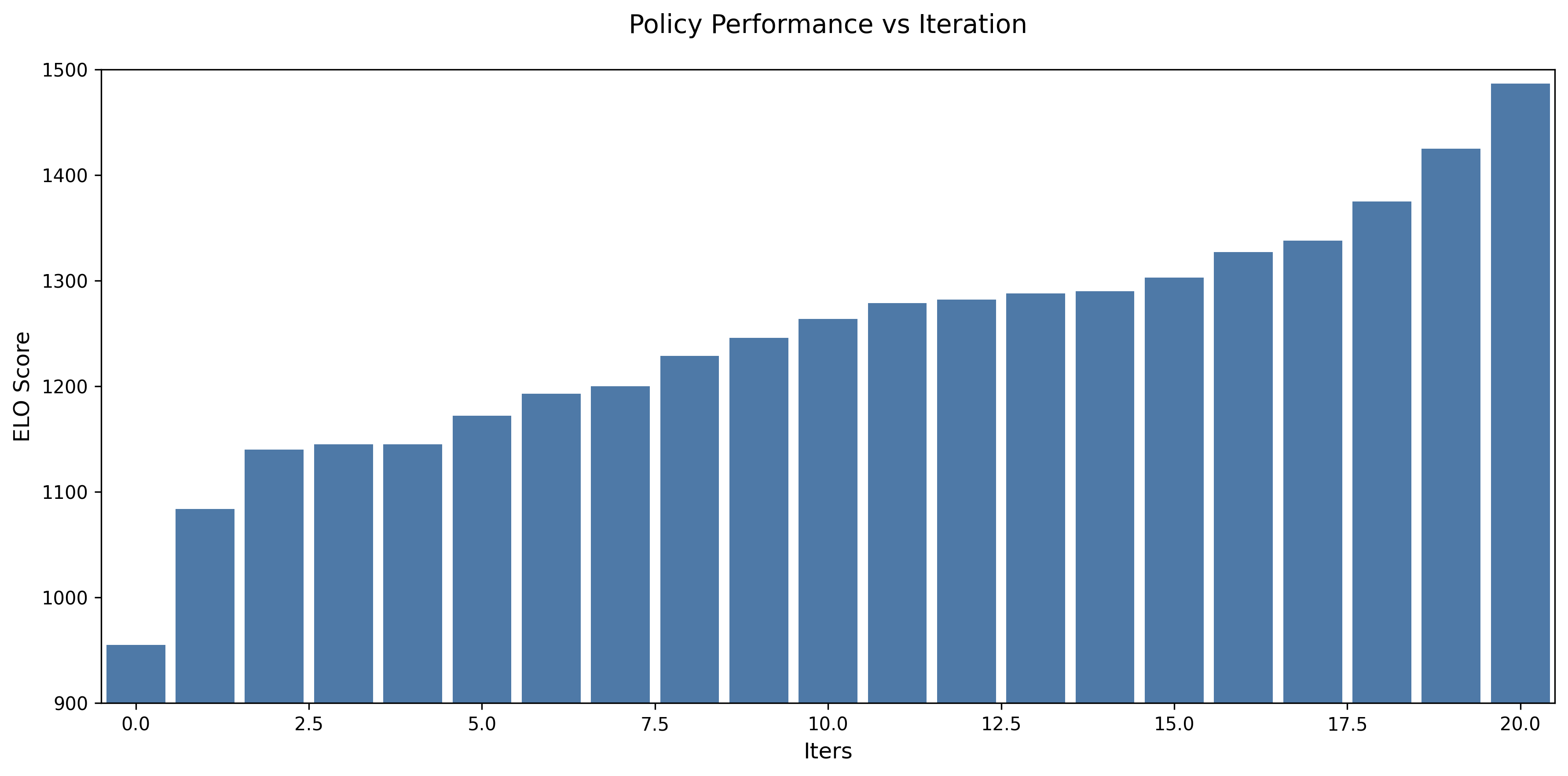}}
\caption{The ELO score gradually improves with policy iteration.}
\label{fig:elovsiter}
\end{figure}

First, we validate that PolicyEvolve can stably improve policies. After $20$ improvements, we pull all policy codes from the Global Pool (including a \textit{Random} policy), have them compete against each other multiple times, and calculate their ELO scores. The results are shown in Fig.~\ref{fig:elovsiter}. We observe that with the continuous improvement of PolicyEvolve, the newer the policy, the higher the ELO score, indicating that the policy quality is indeed stably improving.

\begin{figure}[!ht]
\centering
\centerline{\includegraphics[width=\textwidth]{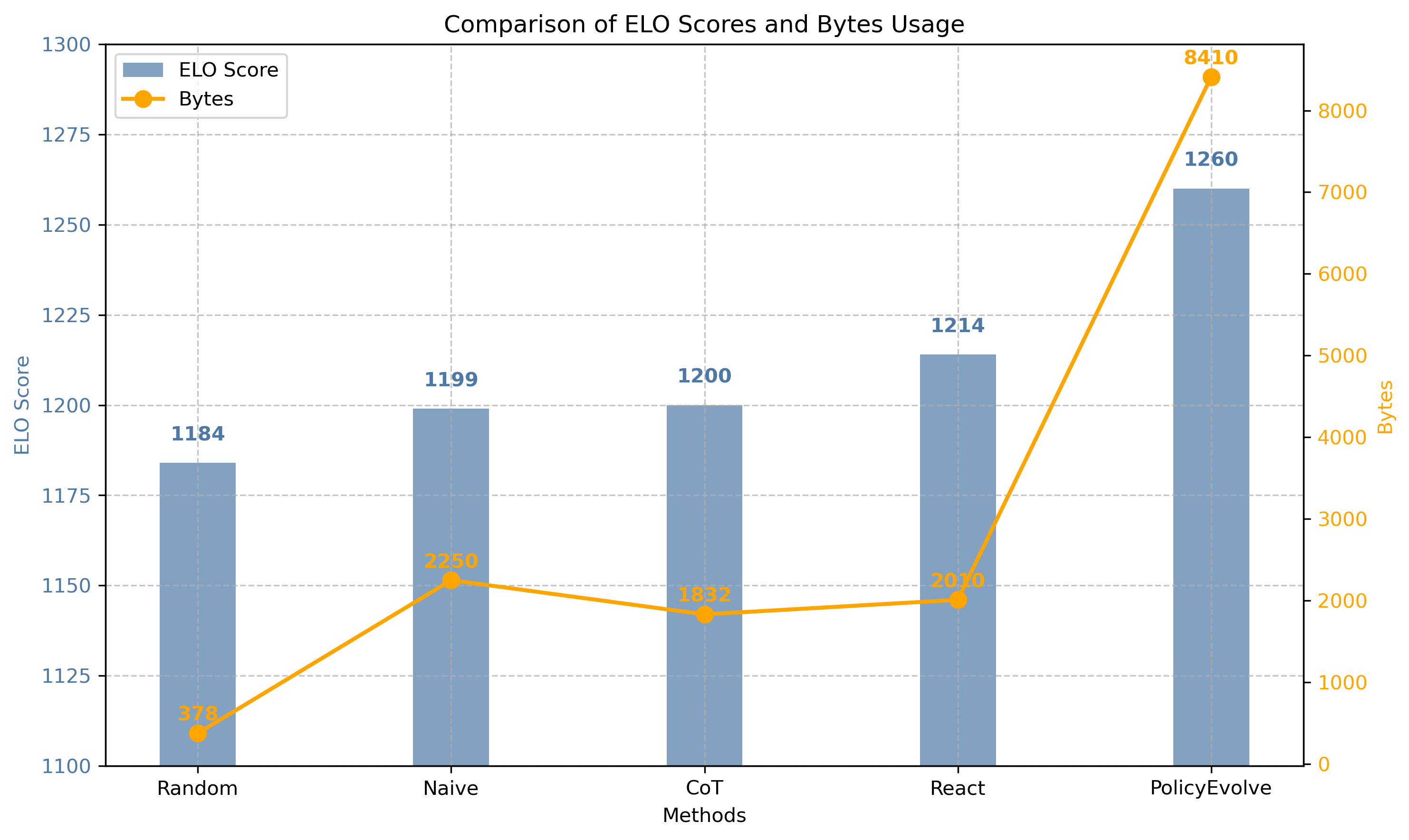}}
\caption{Comparison of five policies in head-to-head battles, showing ELO scores and code sizes.}
\label{fig:elowin}
\end{figure}

We conducted head-to-head battles between the policy trained by PolicyEvolve for $20$ rounds and the code generated by \textit{Random}, \textit{Naive}, \textit{CoT}, and \textit{React}. We placed these five policies into a policy pool and randomly sampled pairs for battles, calculated the ELO score for each policy, and recorded the code size of each policy, as shown in Fig.~\ref{fig:elowin}.

The results demonstrate that the policy generated by our algorithm significantly outperforms others in ELO score, followed by \textit{React} and \textit{CoT}, both of which show moderate improvements in LLM code generation. Meanwhile, due to the relative complexity of our policy, it exhibits a larger code size, indicating that our framework maintains stability when generating complex code and can produce longer algorithms. Furthermore, we compared the pairwise win rates among all five policies, as presented in Tab.~\ref{tab:winrate}. The policy generated by our PolicyEvolve consistently outperforms all baseline policies.

\begin{table*}[ht]
\caption{The win rate between each policy. Each cell represents the win rate of \textit{Column} policy.}
\label{tab:winrate}
\centering
\renewcommand{\arraystretch}{1.2}
\begin{tabular}{lccccc}
\toprule
 & \textbf{Random} & \textbf{Naive} & \textbf{CoT} & \textbf{React} & \textbf{PolicyEvolve}\\
\midrule
\textbf{Random} & - & $0.54$ & $0.92$ & $0.92$ & $1.00$ \\
\textbf{Naive} & - & - & $0.88$ & $0.94$ & $1.00$ \\
\textbf{CoT} & - & - & - & $1.00$ & $1.00$ \\
\textbf{React} & - & - & - & - & $1.00$ \\
\textbf{PolicyEvolve} & - & - & - & - & - \\
\bottomrule
\end{tabular}
\end{table*}

\subsection{Ablation Study}
\begin{figure}[!ht]
\centering
\centerline{\includegraphics[width=\textwidth]{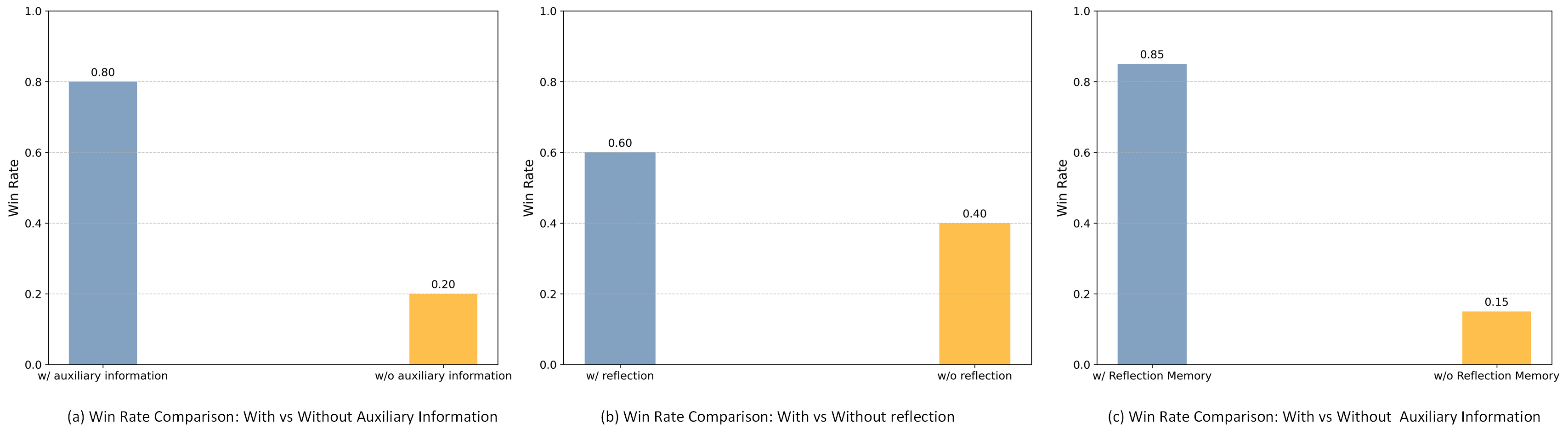}}
\caption{Win Rate Comparison with different prompt tactics}
\label{fig:abs1}
\end{figure}

The prompt for policy generation and iterative improvement significantly impacts policy quality. In the policy generation phase, we noted that auxiliary information beyond environmental descriptions in the prompt, such as providing hints like "boundary colors," enables the initial policy to adhere to game rules. We compared policies with auxiliary information against those without, and their win rates in adversarial gameplay are shown in Fig.~\ref{fig:abs1}(a), demonstrating that policies with auxiliary information achieve significantly higher win rates than those without. Subsequently, we compared policy iteration with and without reflection, as directly feeding experiential data to LLMs to generate improved policies faces challenges of large token volumes and low information density, leading LLMs to hallucinate and overlook critical experiences. Therefore, in our PolicyEvolve framework, we employ a two-step approach: first, we use separate LLMs to summarize experiential data, analyze success/failure causes, and derive improvement suggestions; then, based on these suggestions and historical policy code, we generate refined policies. The win rates of adversarial gameplay between these two approaches are presented in Fig.~\ref{fig:abs1}(b), revealing that the two-step approach markedly outperforms direct generation from experiential data. Finally, we compared the effects of with and without Reflection Memory, which stores reflections and improvement suggestions from historical policy iterations as a long-term memory. The win rates in adversarial gameplay are depicted in Fig.~\ref{fig:abs1}(c), showing that policies with Reflection Memory achieve significantly higher win rates than those without.

\subsection{Toy Example}
To intuitively understand the policy generation process of PolicyEvolve, we designed a toy example. First, we initialized a global pool containing only one random policy (generating actions randomly). PolicyEvolve generated the initial policy $\pi_0$ based on environmental descriptions and task objectives, with the LLM outputting the following policy rationale (code omitted): 

\textit{1. boundary detection takes highest priority to ensure no contact with boundaries while chasing enemies or moving randomly.}

\textit{2. within safe boundaries, locate enemies and calculate force direction and angle based on relative positions, dynamically adjusting force to optimize energy usage.}

\textit{3. if no enemies are detected, enter exploration mode using direction memory for random movement to avoid local wandering and energy waste; energy management is divided into energy-saving and normal modes, stopping actions when energy falls below the reserve threshold to ensure sufficient energy for critical moments.}

\textit{4. action outputs are strictly limited to the ranges [-100, 200] and [-30, 30] to avoid exceeding physical rules.}

Then, $\pi_0$ competed against the random policy in the global pool to gather experiential data for summarization and reflection. The reflection and improvement suggestions output by PolicyEvolve are as follows: 

\textit{Reflections: Briefly summarizing the agent's overall performance and key issues: }

\textit{1. The agent performed adequately in boundary avoidance but over-relied on fixed thrust and angles, leading to local optima in complex environments.}

\textit{2. Enemy tracking was simplistic, lacking multi-enemy scenario handling capabilities, making it vulnerable to encirclement or misdirection.}

\textit{3. The exploration mode’s direction memory showed excessive randomness during prolonged enemy absence, causing energy waste.}

\textit{4. Energy management was basic, failing to account for dynamic environmental changes with overly simplistic energy-saving triggers.}

\textit{Improvement Suggestions: Proposing specific actionable solutions for each error with code examples: }

\textit{1. Boundary detection refinement (detect\_boundaries): Clarify "distance" as actual grid count from agent's observation center to nearest boundary pixel; treat boundaryless sides as safe to align threat assessment with reality rather than display indices.}

\textit{2. Force/angle calculation enhancement (calculate\_force\_and\_angle): Incorporate both (A) alignment between current velocity and target directions and (B) velocity component toward target when determining thrust; allow greater thrust when aligned, prioritize orientation adjustment and reduce thrust during misalignment, with smoothing for continuous outputs.}

\textit{3. Random exploration optimization (explore\_randomly): Introduce inertia by perturbing previous direction slightly instead of complete randomness; permit major direction changes only after sustained steps; reduce exploration intensity and thrust based on energy status to minimize prolonged aimless consumption; implement “recently visited sector” prioritization to improve map coverage.}

\textit{4. Energy strategy improvement (energy judgment in act): Replace fixed-threshold energy saving with tiered strategies (e.g., offensive/balanced/conservative modes) dynamically adjusted based on enemy visibility, distance/count, and recent energy consumption rate; in conservative mode, employ low-energy yet mobile actions (reduced thrust, slow turning) instead of complete shutdown to extend survival while retaining minimal responsiveness.}

PolicyEvolve then generated the improved policy $\pi_1$ based on the improvement suggestions, with the LLM outputting the following policy rationale (code omitted): 

\textit{1. boundary detection takes highest priority to ensure no contact with boundaries while chasing enemies or moving randomly.}

\textit{2. within safe boundaries, locate enemies and calculate force direction and angle based on relative positions.}

\textit{3. if no enemies are detected, enter exploration mode using direction memory for random movement to avoid local wandering.}

\textit{4. energy management is divided into energy-saving and normal modes, stopping actions when energy falls below the threshold to ensure sufficient energy for critical moments. }

Similarly, $\pi_1$ competed against the random policy in the global pool to gather experiential data for summarization and reflection. The win rates of $\pi_0$ and $\pi_1$ against the global pool were $0.47$ and $0.55$ respectively, demonstrating clear policy improvement.

\section{Conclusion and Future Work}
This work introduces PolicyEvolve, the first programmatic reinforcement learning framework designed for multi-agent tasks, which leverages LLMs to autonomously evolve interpretable rule-based policies, addressing critical limitations of traditional MARL including excessive sample dependency, computational inefficiency, and opacity. Its core innovation lies in the dual-pool architecture (Global and Local Pools) trained via Population-Based Training, coupled with an iterative refinement mechanism orchestrated by the Policy Planner and Policy Generator, enabling continuous policy enhancement. Extensive experiments across diverse LLMs and multi-agent tasks demonstrate PolicyEvolve's superior sample efficiency and policy quality compared to prompt-based baselines. Future work will focus on extending the framework to cooperative multi-agent scenarios, developing automated vulnerability detection for policy robustness, exploring hybrid neural-programmatic architectures for high-dimensional state spaces, and optimizing token consumption through adaptive prompting and hierarchical memory summarization to enable scalable deployment.

\bibliographystyle{unsrt}  
\bibliography{references}  


\end{document}